\documentclass[a4paper]{article}
\usepackage{soul}
\usepackage[pages=none, color=black, position={current page.south}, placement=bottom, scale=1, opacity=1, vshift=5mm]{background}
\SetBgContents{~}

\usepackage[margin=1in]{geometry}

\usepackage{amsmath}
\usepackage{amsthm}
\usepackage{amssymb}
\usepackage{booktabs}
\usepackage{subcaption}
\usepackage{soul}
\usepackage{float}
\usepackage[utf8]{inputenc}
\usepackage[
    colorlinks=true,
    linkcolor=RoyalBlue,
    citecolor=RoyalBlue,
    urlcolor=Green
]{hyperref}
\usepackage{doi}

\usepackage[numbers,square]{natbib}

\usepackage{graphicx, color}
\usepackage[dvipsnames]{xcolor}
\graphicspath{{fig/}}

\usepackage{algorithm, algpseudocode}
\usepackage{mathrsfs}

\usepackage{lipsum}

\usepackage{pifont}
\usepackage{wasysym}

\usepackage[dvipsnames]{xcolor}

\newcommand{\dataset}[1]{\textbf{BossouChimpanzee#1}}

\title{\Huge\dataset{}:\\[0.15em]Long-term Chimpanzee Video Dataset}

\author{
Daniel~Schofield\textsuperscript{1,*} \quad
Susana~Carvalho\textsuperscript{2,3,4} \quad
Vladimir~Iashin\textsuperscript{1} \quad
Andrew~Zisserman\textsuperscript{1}\\[0.3em]
Max~Bain\textsuperscript{1} \quad
Arsha~Nagrani\textsuperscript{1} \quad
David~Ng\textsuperscript{1} \quad
Claudia~Sousa\textsuperscript{\dag} \quad
Boniface~Zogbila\textsuperscript{5}\\[0.3em]
Jules~Dor\'e\textsuperscript{5} \quad
Dora~Biro\textsuperscript{6} \quad
Misato~Hayashi\textsuperscript{7,8} \quad
Tetsuro~Matsuzawa\textsuperscript{7,9}
}

\date{
$^1$Visual Geometry Group, Department of Engineering Science, University of Oxford, UK\\
$^2$University of Oxford, UK\\
$^3$CIBIO--BIOPOLIS, University of Porto, Vairão, Portugal\\
$^4$Department of Science, Gorongosa National Park, Mozambique\\
$^5$Institut de Recherche Environnementale de Guinée (IREG), Bossou, Guinea\\
$^6$University of Rochester, USA\\
$^7$Chubu Gakuin University, Japan\\
$^8$Japan Monkey Centre, Japan\\
$^9$Northwest University, China\\[0.5ex]
$^\dag$In memoriam\\[1em]
$^*$Corresponding author: \texttt{dans@robots.ox.ac.uk}
}

\begin{document}
	\maketitle
\vspace{-1em}
\begin{center}
\Large
\color{ForestGreen}
\href{https://robots.ox.ac.uk/\string~vgg/data/BossouChimpanzee/}
     {\nolinkurl{robots.ox.ac.uk/~vgg/data/BossouChimpanzee/}}
\end{center}
\vspace{1em}

	\begin{abstract}
		We describe the \dataset{} video dataset, a unique long-term visual record of wild chimpanzees at an outdoor laboratory for field experiments in Bossou, Guinea, spanning three decades (1988--2018) and comprising over 1,200 hours of continuous video recordings collected through collaborative fieldwork and research. In this paper, we outline the history and scientific contributions of the experimental paradigm and video archive, provide key statistics and details on the structure of the main video dataset, and release an initial $\sim$74h snapshot, \textbf{BossouChimpanzee70h}, covering 23 identified individuals focused on chimpanzee individual and action recognition, ahead of the full video resource. This dataset represents a valuable resource for cognitive and behavioural research in ethology and a rich benchmark for training and evaluating machine learning models on audiovisual data from the wild.
		
	\end{abstract}

\clearpage
	
	\section{Introduction}
	\label{sec:intro}

The \dataset{} video dataset is a large audiovisual archive of wild chimpanzees (\emph{Pan troglodytes verus}) recorded at the long-term field site of Bossou, Guinea, spanning 1988 to 2018. \dataset{} represents a unique, naturalistic dataset, from continuous, long-term, and systematically collected footage of wild chimpanzees over several decades. The videos show a community of wild chimpanzees in a forest clearing, predominantly using stone tools to open nuts. In addition to its duration, the archive is distinguished by its quasi-experimental design: it was established as an `outdoor laboratory' to study how behaviour emerges under controlled ecological conditions, particularly in the context of tool use. By controlling the availability of stone tools and nuts, researchers were able to explore the processes driving behavioural variation in a wild population. Bossou is unique,  not only for the length and consistency of its video record but also for the richness of the behaviour it captures through its experimental paradigm. The site has been central to our understanding of stone tool use and cultural transmission~\cite{matsuzawa_emergence_2008, Biro2003-by}, offering rare opportunities to model the evolution of cognition and intelligence in the wild~\cite{Carvalho2009-wn, carvalho2012chimpanzee}.

\begin{figure}
    \centering
    \includegraphics[width=1\linewidth]{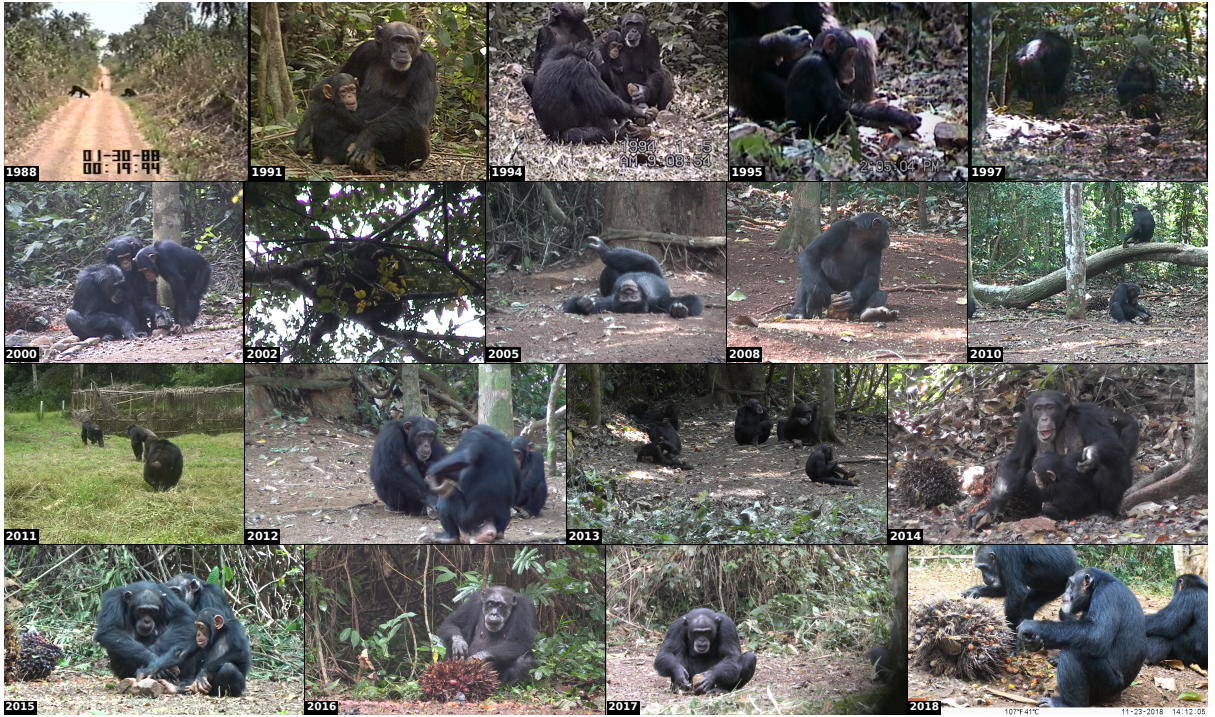}
    \caption{A sample of frames from a selection of years of the \dataset{} dataset (1988--2018).}
    \label{fig:sample_frames}
\end{figure}

We release a slice of the 1,200-hour archive totalling $\sim$74 hours of video (with audio), which we refer to as \dataset{70h}.
This subset contains 226 files and 7.8M frames depicting the behaviour of 23 chimpanzees.
The footage comprises field recordings from six years (2000, 2004, 2006, 2008, 2012, and 2013) and a sample of frames 
is shown in Figure~\ref{fig:sample_frames}.
It is annotated with 7.3M \textit{face} boxes, grouped into 12.6k tracks; each track is assigned one of the 23 identities.
In addition, 2.3M \textit{body} boxes across 3.6k tracks are annotated, covering 13 individuals in 2012--2013.
To study action recognition, nearly 11k \textit{action} annotations are provided across 11 individuals in 2012--2013, including 8.4k linked to body tracks.

With the release of this dataset, our aim is to create a large-scale ecosystem that supports collaborative scientific research and advances the development of machine learning models across diverse disciplines --- from AI to ethology and conservation. We also highlight the importance of recognising the long-term contributions of the researchers and field assistants who helped build this resource, which we hope will benefit the research community for generations. In addition, we emphasise the critical need to protect wild chimpanzee populations and support the communities who coexist with them. 

In the following sections we briefly outline the research context and history of the field site, 
the protocols for data collection, and the structure of the video dataset, 
including accompanying annotations for training computer vision models for several tasks: 
(i) individual recognition; 
(ii) face and body detection and tracking; and (iii) spatio-temporal audiovisual action recognition.
Finally, we outline the implications for behavioural research and machine learning, and future directions and investigations for the dataset.

\section{Dataset Context: The Outdoor Laboratory}
\subsection{History and Scientific Context}

The Bossou field site is located in southeastern Guinea, $\sim$4\,km from the Liberia border and $\sim$6\,km from the Nimba mountain range. The area represents a mosaic habitat of primary and secondary forest \cite{humle2011location}. A community of chimpanzees has been studied here for decades and has coexisted with humans from the village of Bossou, holding deep cultural significance for the local Manon people \cite{Yamakoshi2011-st}.

Long-term research at Bossou began in the 1960s, when Dutch scientists first visited the site \cite{Kortlandt1967-tm, Yamakoshi2011-st}, and was later formalised by Kyoto University. Yukimaru Sugiyama established one of the first long-term field programmes that continuously observed wild chimpanzees (1976--1986), which was later continued and expanded by Tetsuro Matsuzawa. The site has become scientifically important for documenting key aspects of chimpanzee behaviour and culture, including tool use \cite{Sugiyama1979-wc,Humle2011-kk, Sakura1991-gs}, social dynamics \cite{sugiyama_social_1979, sugiyama_population_1984}, and other behaviours such as dead-infant carrying \cite{biro2010chimpanzee} and ethanol ingestion \cite{hockings2015tools}. Over the decades, the population has faced increasing pressures from habitat loss, encroachment, and isolation. In more than forty years, no females have been recorded joining the community, leading to a gradual decline and an ageing population structure \cite{sugiyama_demography_2011, sugiyama_demographic_2004}. Disease outbreaks \cite{humle_2003_2011} have also further reduced numbers, and by 2022, only three chimpanzees remained after the deaths and disappearances of most of the group.

\subsection{Field Experimental Paradigm}

A central focus of research at Bossou has been the stone nut-cracking culture --- one of the most complex forms of tool use observed in wild chimpanzees \cite{biro2006ontogeny, matsuzawa1996chimpanzee}. Compared with other populations that exhibit the behaviour, chimpanzees at Bossou use a unique combination of stone tools, combining a mobile hammer and anvil in a complex sequence of actions to crack open oil palm nuts (\emph{Elaeis guineensis}), with individuals exhibiting selectivity in both tool choice and technique \cite{Carvalho2008-pb, Carvalho2009-wn}. Numerous studies have investigated how this behaviour is acquired and socially transmitted, with young individuals selectively observing proficient individuals during peering events, a process described as `education by master-apprenticeship' \cite{matsuzawa_emergence_2008, Biro2003-by}. Bossou provides a uniquely long and systematically recorded video archive of this behaviour, providing an opportunity to examine the social dynamics \cite{sugiyama_grooming_1988}, ontogeny \cite{Inoue-Nakamura1997-pq, biro2006ontogeny}, and evolution of tool use \cite{Matsuzawa2011-kd}. As such, the site provides an important model for understanding the origins of technological intelligence and cultural learning in our own evolutionary lineage \cite{Carvalho2009-wn,carvalho2012chimpanzee}.

Field experiments --- where researchers control certain variables under quasi-experimental conditions while observing individuals in natural habitats --- have revolutionised ethology \cite{Matsuzawa2011-kd}. Tetsuro Matsuzawa, influenced by research on chimpanzee cognition in the laboratory, pioneered field experiments on tool use at Bossou \cite{Matsuzawa1994-az}. 

To document this behaviour systematically, researchers provisioned stones and nuts, both naturally available in the surrounding habitat, creating a semi-structured experimental setting known as the outdoor laboratory. Using video cameras mounted on tripods, researchers recorded chimpanzees in a forest clearing as they cracked open naturally available nuts with stone tools. The primary site, known as the `Bureau' ($07^\circ38'41.5''$\,N; $008^\circ29'50.0''$\,W) is a clearing of $\sim$7\,$\times$\,20\,m established in 1991. A second site, known as the `Salon' ($07^\circ38'39.2''$\,N; $008^\circ29'58.5''$\,W) was later established in 2009 on Mount Guein to facilitate closer observation of individuals frequenting that area \cite{Matsuzawa2011-kd}. Across years, different nut species were provisioned depending on seasonal availability, and a matrix of stone tools were numbered so the material properties and selectivity by the chimpanzees could be recorded \cite{Matsuzawa2011-kd}. These outdoor lab experiments were conducted annually between 1986 and 2018, with a few years in which data collection and field research was not possible at the site. 

\subsection{Recording Protocol}
The outdoor laboratory was monitored using one or more video cameras on fixed tripods \cite{Matsuzawa2011-kd}. Depending on the year, recordings were made with a single camera or multiple cameras capturing both wide-angle and zoomed views. Each session began when the first chimpanzee entered the site and ended when the last individual departed. For every session, a PAB (`Presence at Bureau') or PAS (`Presence at Salon') record was kept, manually logging the identity and timing of each chimpanzee present for each experimental session. 
The resulting audiovisual data capture not only tool-use behaviour, such as the choice of stone tools, selection of nuts, and variation among multiple nut-crackers, but also a range of accompanying behaviours, including tool use for drinking water from tree hollows, vocalising, climbing, play, aggression, and affiliative interactions such as grooming. These recordings form one of the most detailed long-term visual archives of naturally occurring tool use and associated social behaviour in wild chimpanzees. In recent years, the archive has supported both the development of machine learning methods~\cite{schofield2019chimpanzee, bain2019count, bain2021automated,schofield2023automated, iashin2025self} and studies of chimpanzee behaviour and cognition~\cite{mielke2022chimpanzee, berdugo2024ontogeny, howard2024nonadjacent, howard2025old, braun2025stone}.

\subsection{Source Collection and Digitisation (1988--2018)}

The full \dataset{} video archive spans recordings from 1988 to 2018 and represents the raw foundation from which an annotated subset \dataset{70h} (see Section~\ref{sec:70h}) was derived. 
The majority of systematic filming took place between 1993 and 2012 (see Table~\ref{tab:dist2_years} for yearly distribution).

The archive comprises $\sim$1,200 analogue tapes which were recorded across Hi8, MiniDV, Betacam, and VHS formats, 
corresponding to roughly 800 unique video files once duplicates were removed. 
When combined with later digital recordings collected between 2012 and 2018, the dataset contains 1,491 video files and over 1,200 hours of footage in total (Table~\ref{tab:bc_full_stats}).

Most footage was filmed using Sony Hi8 and MiniDV cameras, often capturing multiple camera angles per session to record continuous sequences of activity at the outdoor laboratory. Each directory in the archive corresponds to a camera view and session, preserving the original temporal continuity of the recordings.

Video quality varies across years and formats: older footage typically has a resolution of 720 $\times$ 480, 
while later digital recordings are of higher resolution and fidelity. 
Some analogue tapes show minor degradation or signal loss due to age and environmental conditions. A detailed breakdown of video formats, resolutions, and durations is provided in Table~\ref{tab:bc_full_stats}.

\section{\dataset{70h}}\label{sec:70h}
    
    This section outlines a $\sim$74-hour slice of the full video archive, comprising 226 videos spanning six years (2000, 2004, 2006, 2008, 2012, and 2013).
    To unify encoding across all files, the digitised tapes and initially-digital videos were transcoded using \texttt{H.264} and \texttt{AAC} codecs for video and audio streams, preserving the original frame rates, resolution, and perceptual quality. The distribution of video properties is described in Table~\ref{tab:bc70h_stats}.

    This subset provides annotations for face- and body-level detection, tracking, and individual identification, together with frame-level identity labels and audiovisual action labels, each described below.

    \begin{table}[t]
    \centering
    \small
    \begin{tabular}{rrrrr|rrrr|rrr}
    \toprule
    \multicolumn{1}{c}{} & \multicolumn{4}{c|}{Face} & \multicolumn{4}{c|}{Body} & \multicolumn{3}{c}{Action} \\
    \cmidrule(lr){2-5}\cmidrule(lr){6-9}\cmidrule(lr){10-12}
    Year & Boxes  & Videos & Tracks & IDs & Boxes & Videos & Tracks & IDs & Videos & Tracks & IDs \\
       \midrule
    2000 & 1,666,984 & 46 & 2,504 & 21 & -- & -- & --   & -- & -- & --  & -- \\
    2004 & 1,436,429 & 23 & 1,877 & 12 & -- & -- & --   & -- & -- & --  & -- \\
    2006 &   931,924 & 20 & 1,924 & 11 & -- & -- & --   & -- & -- & --  & -- \\
    2008 & 1,607,059 & 49 & 2,819 & 13 & -- & -- & --   & -- & -- & --  & -- \\
    2012 & 1,419,441 & 18 & 2,973 & 13 & 1,878,291& 17 & 2,674 & 13 & 11 & 536 & 8 \\
    2013 &   243,265 &  9 &   463 &  9 &   469,425& 19 &   927 &  9 & 15 & 157 & 8 \\
    \midrule
    All  & 7,305,102 & 165& 12,560& 23 & 2,347,716& 36 & 3,601 & 13 & 26 & 693 & 11\\
    \bottomrule
    \end{tabular}
    \caption{\dataset{70h}: Year-wise summary. Combined per-year counts across three identity annotation levels. 
    For \emph{Face} and \emph{Body} we report the number of videos, tracks, bounding boxes, and individuals (IDs) that appear in a year. 
    The \emph{Action} columns give videos, tracks, and individuals with additional annotations for spatio-temporal (body-level) action recognition (nut-cracking).
    The \emph{Action} \emph{Tracks} column counts tracks from the earlier body annotation set
    (Section~\ref{sec:body_annots}), not the \emph{Body} tracks reported in the same row.}
    \label{tab:per_year_summary}
    \end{table}

\begin{figure}[h]
    \centering
    \includegraphics[width=1\linewidth]{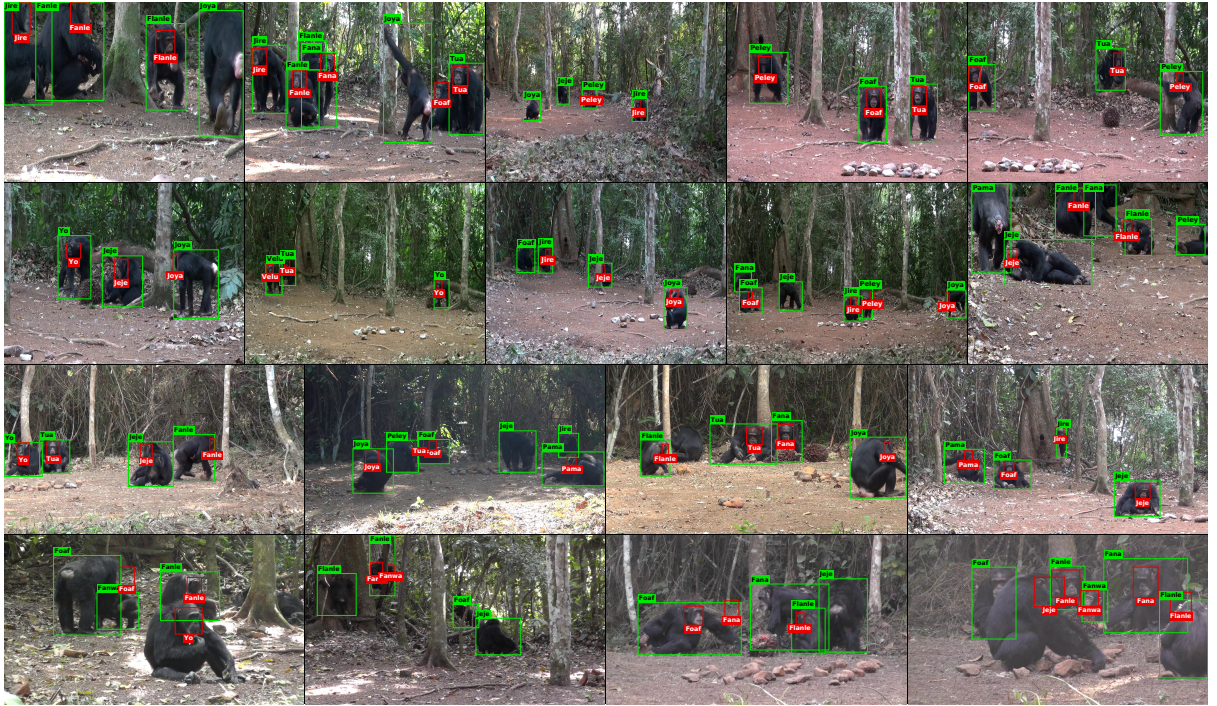}
    \caption{A sample of frames from \dataset{70h} showing \textit{Body} and \textit{Face} annotations. Green boxes show \textit{Body} annotations (SAM~3 tracks~\cite{carion2025sam3segmentconcepts}; Section~\ref{sec:body_annots}) and red boxes show \textit{Face} annotations (\cite{schofield2019chimpanzee,bain2019count}; Section~\ref{sec:face_annots}); text labels indicate identity.}
    \label{fig:body_face_annots}
\end{figure}
\begin{table}[t]
    \centering
    \small
    \setlength{\tabcolsep}{4.0pt}
    \begin{tabular}{l r r r r | r r r r | r r r r}
    \toprule
    & \multicolumn{4}{c}{Face} & \multicolumn{4}{c}{Body} & \multicolumn{4}{c}{Action} \\
    \cmidrule(lr){2-5} \cmidrule(lr){6-9} \cmidrule(lr){10-13}
    Individual               & Boxes  & Vids & Tracks & Yrs & Boxes & Vids & Tracks & Yrs & Actions & Vids & Tracks & Yrs \\
    \midrule
    Jeje (M, 1997)           & 1,154,000 & 113    & 1,803    & 6     & 384,289 & 29     & 355      & 2     & 2,287 & 16     & 87     & 2    \\
    Jire (F, 1958)           & 860,987   & 80     & 1,056    & 6     & 218,196 & 21     & 281      & 2     & 722   & 10     & 66     & 2    \\
    Fanle (F, 1997)          & 670,403   & 68     & 1,201    & 6     & 248,469 & 23     & 569      & 2     & 2,043 & 13     & 201    & 2    \\
    Peley (M, 1998)          & 635,376   & 81     & 1,182    & 5     & 220,987 & 11     & 239      & 1     & 1,111 & 9      & 33     & 1    \\
    Yolo (M, 1991)           & 593,501   & 82     & 683      & 4     & --      & --     & --       & --    & --    & --     & --     & --   \\
    Foaf (M, 1980)           & 531,407   & 79     & 828      & 6     & 253,360 & 23     & 348      & 2     & 1,080 & 12     & 82     & 2    \\
    Yo (F, 1961)             & 439,728   & 62     & 488      & 6     & 93,458  & 8      & 48       & 2     & 192   & 3      & 5      & 1    \\
    Fana (F, 1956)           & 390,616   & 53     & 484      & 6     & 81,024  & 14     & 164      & 2     & 117   & 1      & 11     & 1    \\
    Tua (M, 1957)            & 379,456   & 57     & 570      & 5     & 135,588 & 8      & 104      & 1     & 1,136 & 4      & 64     & 1    \\
    Joya (F, 2004)           & 362,966   & 65     & 1,565    & 4     & 270,958 & 11     & 374      & 1     & 1,262 & 9      & 104    & 1    \\
    Flanle (M, 2007)         & 274,413   & 33     & 842      & 4     & 252,682 & 21     & 671      & 2     & 355   & 8      & 36     & 2    \\
    Velu (F, 1959)           & 222,180   & 29     & 243      & 5     & 36,515  & 4      & 20       & 2     & 73    & 1      & 4      & 1    \\
    Pama (F, 1967)           & 205,875   & 34     & 326      & 5     & 85,126  & 7      & 95       & 1     & --    & --     & --     & --   \\
    Fotaiu (F, 1991)         & 150,089   & 21     & 247      & 1     & --      & --     & --       & --    & --    & --     & --     & --   \\
    Kai (F, 1950)            & 101,996   & 15     & 147      & 1     & --      & --     & --       & --    & --    & --     & --     & --   \\
    Vuavua (F, 1991)         & 91,636    & 10     & 127      & 1     & --      & --     & --       & --    & --    & --     & --     & --   \\
    Juru (F, 1993)           & 65,172    & 21     & 201      & 1     & --      & --     & --       & --    & --    & --     & --     & --   \\
    Fanwa (M, 2012)          & 63,068    & 14     & 307      & 2     & 36,239  & 17     & 233      & 2     & --    & --     & --     & --   \\
    Nto (F, 1993)            & 47,966    & 12     & 77       & 1     & --      & --     & --       & --    & --    & --     & --     & --   \\
    Pili (F, 1987)           & 23,526    & 7      & 62       & 1     & --      & --     & --       & --    & --    & --     & --     & --   \\
    Nina (F, 1954)           & 21,605    & 10     & 42       & 1     & --      & --     & --       & --    & --    & --     & --     & --   \\
    Pokuru (M, 1996)         & 18,961    & 7      & 78       & 1     & --      & --     & --       & --    & --    & --     & --     & --   \\
    Poni (M, 1993)           & 175       & 1      & 1        & 1     & --      & --     & --       & --    & --    & --     & --     & --   \\
    Unknown                  & --        & --     & --       & --    & 30,825  & 17     & 100      & 2     & --    & --     & --     & --   \\
    \bottomrule
    \end{tabular}
    \caption{\dataset{70h}: Individual-wise summary with \emph{Face}, \emph{Body} annotation levels, and annotations for spatio-temporal (body-level) \textit{Action} recognition (nut-cracking). 
    We report the name, sex Male (\textit{M}) or Female (\textit{F}), year of birth, number of annotated bounding boxes, videos, tracks, and years an individual appears in, and the number of action annotations. Body annotations were re-generated using a SAM~3~\cite{carion2025sam3segmentconcepts} + ByteTrack~\cite{zhang2022bytetrack} pipeline; \emph{Unknown} denotes an on-screen chimpanzee whose identity could not be verified.
    \emph{Actions} counts action annotations attributed to a named individual; a further 517
    annotations for which the actor could not be identified are not included.
    The \emph{Action} \emph{Tracks} column counts tracks from the earlier body annotation set
    (Section~\ref{sec:body_annots}), not the \emph{Body} tracks reported in the same row. }
    \label{tab:per_individual_summary}
    \end{table}

    \subsection{Face Annotations}\label{sec:face_annots}
    
    The face annotations were introduced in Schofield~\textit{et al.}~\cite{schofield2019chimpanzee}. 
    First, 3,707 frames sampled every 10 seconds from 2008 footage were manually labelled using the VIA~\cite{dutta2019vgg} web tool
    by drawing tight head bounding boxes (yielding 5,570 boxes), 
    which were used to train an SSD-based~\cite{liu2016ssd} face detector. 
    The trained detector was then run at 25 fps over the archive to produce frame-level face detections, 
    which were grouped into face tracks using a KLT tracker \cite{lucas1981iterative,tomasi1991detection,shi1994good}, 
    with short tracks ($< 10$ frames) pruned as likely false positives. 
    Next, VIA was used to show the first/middle/last frame of each track so a domain expert annotator could assign a single chimpanzee 
    identity label per track, enabling millions of faces to be labelled with only hours of effort. 
    In total, 19,512 tracks from the years 2000, 2004, 2008, and 2012 across 24 labels 
    were labelled (23 individuals and 1 for faulty tracks to be discarded). 
    An example of released face annotations is shown in Figure~\ref{fig:body_face_annots} and
    a summary of the released annotations per year is provided in Table~\ref{tab:per_year_summary} and per individual in Table~\ref{tab:per_individual_summary}.

    Schofield~\textit{et al.}~\cite{schofield2019chimpanzee} used these annotations to train a supervised face recognition model
    and to conduct a social network study~\cite{schofield2023automated}.
    This was followed by Bain~\textit{et al.}~\cite{bain2019count} who used a subset of years (2012 and 2013) and 
    13 individuals (see Table~\ref{tab:per_individual_summary}) to train a face recognition model.
    Iashin~\textit{et al.}~\cite{iashin2025self} used a 338k-box subset covering 14 individuals as a part of the training mixture for a self-supervised face recognition model for chimpanzees; the remaining nine individuals were used for evaluation (see Table~\ref{tab:per_individual_summary}).

    \subsection{Body Annotations}\label{sec:body_annots}

    Body annotations for the Bossou chimpanzees were introduced by Bain~\textit{et al.}~\cite{bain2019count,bain2021automated}. The body labelling pipeline largely mirrors the face bounding-box pipeline described in Section~\ref{sec:face_annots}. 
    First, $\sim$16k body bounding boxes were manually annotated and used to train an SSD-based body detector; the detector was run over the archive and its detections linked into tracks with the `Siamese-RPN' tracker~\cite{li2018high}, and a domain expert annotator used VIA to assign a single chimpanzee identity to each track. The set was later expanded to its final version in Bain~\textit{et al.}~\cite{bain2021automated}, where it was used to visually
    ground the scene-level nut-cracking action annotations described in Section~\ref{sec:action_annots}, which were then used to train a spatio-temporal
    action recognition model.
    For this release, we used a three-stage approach (by David Ng) to reprocess the original footage to mitigate the bounding box inaccuracy and drift sometimes seen in previously released annotations. First, the SAM~3 detector~\cite{carion2025sam3segmentconcepts} was used with the text prompt \texttt{chimpanzee} and a confidence threshold of 0.9 to detect chimpanzee bodies in each frame; these detections were linked into tracks with ByteTrack~\cite{zhang2022bytetrack}, producing high-quality chimpanzee body track proposals without identity labels. Second, identity labels from the original annotations were carried over using an automated assignment stage based on each proposed track's accumulated Intersection over Union (IoU) overlap with tracks in the original annotations. Each track was assigned an identity transfer confidence level based on the dominance of the accumulated IoU overlap. Third, short tracks ($< 20$ frames) were discarded and a human annotator examined the subset of proposals with high-confidence identity transfers and verified their correctness using a customised LISA web annotation tool \cite{dutta2019vgg}. Proposals whose identity transfer was ambiguous or unsupported were retained as tracks but labelled \emph{Unknown} rather than assigned an identity. This pipeline produced longer, less fragmented tracks with higher fidelity. The final set of body annotations comprises $\sim$3.6k tracks and 2.3M body instances across 13 individuals (Table~\ref{tab:per_year_summary}).
    An example of released body annotations is shown in Figure~\ref{fig:body_face_annots} and summaries are provided per year in Table~\ref{tab:per_year_summary} and per individual in Table~\ref{tab:per_individual_summary}. The earlier tracks of Bain~\textit{et al.}~\cite{bain2021automated} remain available and are distributed with the action annotations (Section~\ref{sec:action_annots}), which are currently linked to them rather than to the SAM~3 tracks described above.

    \subsection{Frame Annotations}
    We release the frame-level identity annotations introduced by
    Bain~\textit{et al.}~\cite{bain2019count}, which enumerate all chimpanzee
    identities visible in a video frame. 
    These labels were produced manually and are not derived from finer-grained annotations. 
    They provide an additional source of individual grounding beyond the semi-manual annotations 
    used for the body tracks in Bain~\textit{et al.}~\cite{bain2019count}.
    In total, we release 0.9M annotated frames from 27 videos from 2012 and 2013,
    and further frame-level annotations could be mined from other years and videos
    that have face- and body-level box identity annotations.
    
    In \cite{bain2019count}, the frame-level labels were used to train a
    chimpanzee counting model, which was then used to train a body-level
    chimpanzee recognition model without relying on body tracks.
    
    \begin{table}[t]
    \centering
    \small
    \setlength{\tabcolsep}{6pt}
    \begin{tabular}{l c c c c c}
    \toprule
     & Bouts & Strikes & Failures & Successes & Total \\
    \midrule
    Count & 1,281 & 7,884 (7,367) & 588 & 1,142 & 10,895 \\
    \bottomrule
    \end{tabular}
    \caption{\dataset{70h}: Action (nut-cracking) fine-grained annotation counts.
    The strike count includes 7,367 strikes by identified individuals (in parentheses), and
    517 `off-screen' strikes where an actor could not be seen.}
    \label{tab:action_types}
    \end{table}

    \begin{figure}[t]
        \centering
        \includegraphics[width=1\linewidth]{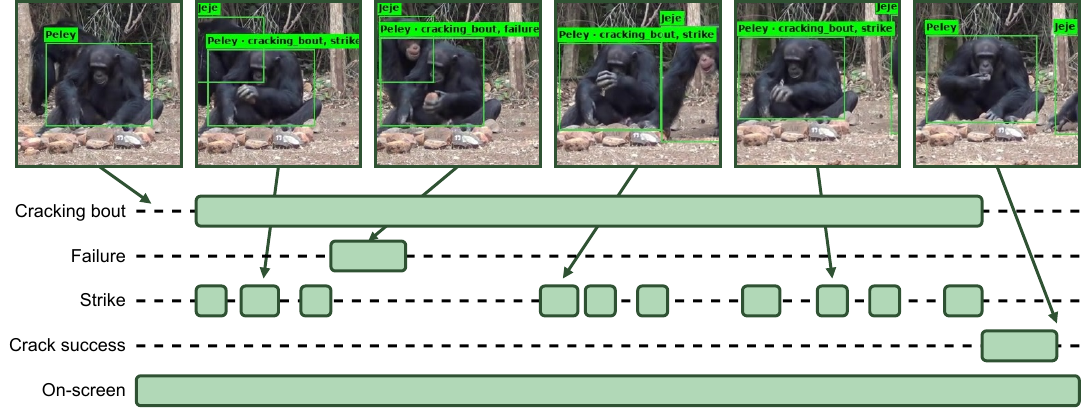}
        \caption{
            \textbf{Example of an on-screen nut-cracking action annotation.}
            An on-screen individual, Peley, performs a successful cracking bout comprising 10 strikes.
            The bout includes one failure when the nut slips away. Another individual, Jeje, passes
            behind him.
        }
        \label{fig:nut_crack_annot}
    \end{figure}

    \subsection{Audiovisual Spatio-Temporal Action Annotations}\label{sec:action_annots}

    We release annotations of nut-cracking behaviour for 2012 and 2013, comprising on-screen
    (spatio-temporal) and off-screen (scene-level, audio-only) labels, including `striking' (hitting a nut with a stone tool), `failure' (the strike misses or a nut slips after a hit), and a `cracking bout' (all instances of striking and failure up to `crack success', the moment the nut shell breaks), followed by a chimpanzee eating the exposed kernel. A detailed description is provided on the website. 
    The annotations were produced via the iterative process described below.
    
    First, the body boxes for 2012 and 2013 from~\cite{bain2019count} (Section~\ref{sec:body_annots}) were used to seed
    the initial temporal nut-cracking labels and to filter out scenes without a chimpanzee in shot. 
    Next, striking segments from the initial labels were used to fine-tune an audio-only
    ResNet-18~\cite{he2016deep} pretrained on VGGSound~\cite{chen2020vggsound}. 
    This model generated temporal striking proposals for unannotated 2012--2013 footage, covering both on-screen and
    off-screen nut-cracking. 
    Proposals were manually verified and, when necessary, adjusted by an expert
    anthropologist. Finally, track IDs were assigned to on-screen instances. 
    In total, the dataset contains 10{,}895 action annotations, including off-screen striking.
    Annotation details are shown in
    Tables~\ref{tab:per_year_summary}, \ref{tab:per_individual_summary}, and \ref{tab:action_types}. An
    example of an on-screen nut-cracking bout, including striking, failure, and crack success, for one individual is
    provided in Figure~\ref{fig:nut_crack_annot}. These audiovisual spatio-temporal annotations were used by Bain~\textit{et al.}~\cite{bain2021automated} to train an audiovisual action recognition model.

    \section{Future work and \dataset{}}
      We release \dataset{70h} as an initial subset of the wider \dataset{} archive, covering the six years on which most existing annotations focus. Substantial portions of the archive, particularly the earlier recordings, therefore remain unexplored.
      The video archive presents significant opportunities for further annotation and benchmarking across a wide range of 
      computer vision and ethological challenges, including individual identification, pose estimation, action and interaction recognition, and multimodal audiovisual analysis. This offers opportunities to explore questions about long-term behavioural change, ontogeny, learning,  cultural transmission, individual life histories, and the effects of ageing, sociality, and tool use. We plan to release the entire archive once the required content filtering is complete. We hope this resource will stimulate multidisciplinary research collaborations and encourage other research teams to make their data accessible.

	\label{sec:prev-results}

	\paragraph{Acknowledgments.}

 This video dataset is released as a collaborative effort honouring over 50 years of research at the Bossou Field Site and is dedicated to the humans and chimpanzees whose shared history in Guinea has contributed to a better future for conservation and nature. We are grateful to all researchers, field guides, and funders who have contributed to the collection and curation of the data. The video data for this research were provided by Tetsuro Matsuzawa and the Primate Research Institute, Kyoto University, Japan, and we thank the Guinean authorities who granted permission for long-term research (MESRSI, DGRSI, IREG). We also thank the Visual Geometry Group, Department of Engineering Science, University of Oxford, for hosting the dataset and providing infrastructure support. A full list of contributors is available on the project website: \url{https://robots.ox.ac.uk/~vgg/data/BossouChimpanzee/}. We encourage you to donate via the website to help protect wild chimpanzees and support local communities in Guinea. The work for this dataset release was partially funded by the EPSRC Programme Grant VisualAI (EP/T028572/1) and is supported by Schmidt Sciences, LLC.
	
	\bibliography{refs}

\clearpage

	\appendix
	
\section{Media Metadata}

\label{app:metadata}

\renewcommand{\thetable}{A\arabic{table}}

\setcounter{table}{0}

\subsection{\dataset{} Video Archive}

\begin{table}[H]
  \centering
  \captionsetup[sub]{labelformat=empty}

  \begin{minipage}[t]{0.32\textwidth}
    \subcaptionbox{(a) Videos \& Duration per Year\label{tab:dist2_years}}[\linewidth]{%
      \begin{tabular}{r rr r}
        \toprule
        Year & Count & Hours & Share \\
        \midrule
        1988 &  28 &  13.9 &  1.1\,\% \\
        1990 &  13 &  24.0 &  1.9\,\% \\
        1991 &  47 &  67.3 &  5.3\,\% \\
        1992 &  18 &  37.3 &  2.9\,\% \\
        1993 &  32 &  55.6 &  4.4\,\% \\
        1994 &   7 &  14.2 &  1.1\,\% \\
        1995 &  18 &  41.2 &  3.3\,\% \\
        1996 &  48 &  98.6 &  7.8\,\% \\
        1997 &   4 &   5.2 &  0.4\,\% \\
        1998 &  56 &  56.8 &  4.5\,\% \\
        1999 &   6 &  13.0 &  1.0\,\% \\
        2000 &  23 &  22.2 &  1.8\,\% \\
        2002 & 176 & 231.5 & 18.3\,\% \\
        2003 & 156 & 187.7 & 14.8\,\% \\
        2004 &  70 &  91.8 &  7.2\,\% \\
        2005 &  55 &  61.3 &  4.8\,\% \\
        2007 &   8 &  10.3 &  0.8\,\% \\
        2008 &  60 &  78.2 &  6.2\,\% \\
        2009 &  12 &  12.9 &  1.0\,\% \\
        2010 &  32 &  38.5 &  3.0\,\% \\
        2011 &   8 &   7.6 &  0.6\,\% \\
        2012 & 121 &  41.5 &  3.3\,\% \\
        2013 & 217 &  12.1 &  1.0\,\% \\
        2014 &  55 &   4.1 &  0.3\,\% \\
        2015 &  55 &   3.3 &  0.3\,\% \\
        2016 &  43 &  24.7 &  2.0\,\% \\
        2017 &   9 &   4.7 &  0.4\,\% \\
        2018 & 114 &   6.9 &  0.5\,\% \\
        \midrule
        All  & 1491& 1266.5& 100.0\% \\
        \bottomrule
      \end{tabular}
    }%
  \end{minipage}
  \hfill
  \begin{minipage}[t]{0.32\textwidth}
    \subcaptionbox{(b) Video Resolution Distribution\label{tab:dist2_vid_res}}[\linewidth]{%
      \begin{tabular}{crr}
        \toprule
        Resolution & Count & Share \\
        \midrule
        720 $\times$ 480   & 852 & 57.1\,\% \\
        960 $\times$ 540   & 272 & 18.2\,\% \\
        1920 $\times$ 1080 & 152 & 10.2\,\% \\
        1280 $\times$ 720~\,& 126 &  8.5\,\% \\
        1440 $\times$ 1080 &  49 &  3.3\,\% \\
        720 $\times$ 576   &  13 &  0.9\,\% \\
        714 $\times$ 480   &  11 &  0.7\,\% \\
        3840 $\times$ 2160 &   8 &  0.5\,\% \\
        712 $\times$ 480   &   5 &  0.3\,\% \\
        640 $\times$ 480   &   2 &  0.1\,\% \\
        716 $\times$ 480   &   1 &  0.1\,\% \\
        \bottomrule
      \end{tabular}
    }%
    \vspace{2.2ex}
    \subcaptionbox{(c) Audio Sample Rate Distribution\label{tab:dist2_sr}}[\linewidth]{%
      \begin{tabular}{rrr}
        \toprule
        Rate (Hz) & Count & Share \\
        \midrule
        48000  & 761 & 51.0\,\% \\
        32000  & 618 & 41.4\,\% \\
        44100  &  75 &  5.0\,\% \\
         8000  &  37 &  2.5\,\% \\
        \bottomrule
      \end{tabular}
    }%
    \vspace{2.2ex}
    \subcaptionbox{(d) Camera Type Distribution\label{tab:dist2_camera_type}}[\linewidth]{%
      \begin{tabular}{lrr}
        \toprule
        Type & Count & Share \\
        \midrule
        MiniDV     & 640 & 43.7\,\% \\
        \textit{Digital} & \textit{608} & \textit{41.6\,\%} \\
        Hi8     & 152 & 10.4\,\% \\
        VHS     &  33 &  2.3\,\% \\
        DVD     &  26 &  1.8\,\% \\
        Betacam &   4 &  0.3\,\% \\
        \bottomrule
      \end{tabular}
    }%
  \end{minipage}
  \hfill
  \begin{minipage}[t]{0.29\textwidth}
    \subcaptionbox{(e) Duration per Year (chart)\label{fig:dist2_duration}}[\linewidth]{%
      \centering
      \includegraphics[width=\linewidth]{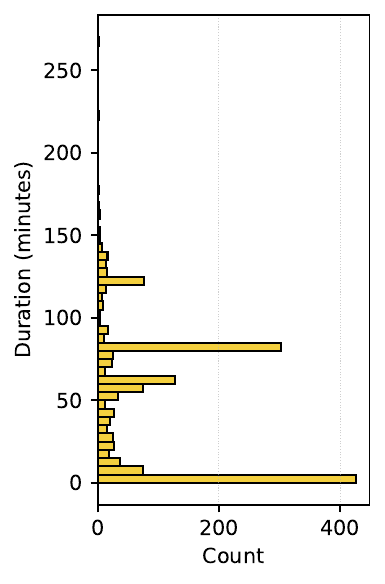}
    }%
    \vspace{0.4ex}
    \subcaptionbox{(f) Video FPS Distribution\label{tab:dist2_fps}}[\linewidth]{%
      \begin{tabular}{rrr}
        \toprule
        FPS & Count & Share \\
        \midrule
        29.97 & 875 & 58.7\,\% \\
        25.00 & 364 & 24.4\,\% \\
        59.94 & 139 &  9.3\,\% \\
        30.00 & 113 &  7.6\,\% \\
        \bottomrule
      \end{tabular}
    }%
    \vspace{2ex}
    \subcaptionbox{(g) Audio Channels Distribution\label{tab:dist2_ch}}[\linewidth]{%
      \begin{tabular}{crr}
        \toprule
        Channels & Count & Share \\
        \midrule
        2 & 1363 & 91.4\,\% \\
        1 &   82 &  5.5\,\% \\
        6 &   46 &  3.1\,\% \\
        \bottomrule
      \end{tabular}
    }%
  \end{minipage}
  \caption{\dataset{}: Media attribute distributions.
  \emph{Count} refers to individual digitised video files, each corresponding to a single archived tape.
  \dataset{70h} (Table~\ref{tab:bc70h_stats}) instead counts clips excerpted from these recordings, so its
  per-year count may exceed the archive count while total duration remains a strict subset.
  \emph{Year} denotes the field season a recording set is archived under, given by the date of its first video.
  Seasons spanning a new year (e.g.\ 2005--2006) are filed under the starting year; Table~\ref{tab:bc70h_stats}
  assigns years by recording date, where such a season appears under the later year.}
  \label{tab:bc_full_stats}
\end{table}

\subsection{\dataset{70h}}

\begin{table}[H]
  \centering
  \captionsetup[sub]{labelformat=empty}
  \begin{minipage}[t]{0.34\textwidth}
    \subcaptionbox{(a) Video Resolution Distribution\label{tab:dist_vid_res}}[\linewidth]{%
      \begin{tabular}{crr}
        \toprule
        Resolution & Count & Share \\
        \midrule
        720 $\times$ 480   & 171 & 75.7\,\% \\
        1440 $\times$ 1080 &  22 &  9.7\,\% \\
        960 $\times$ 540   &  20 &  8.8\,\% \\
        1920 $\times$ 1080 &  13 &  5.8\,\% \\
        \bottomrule
      \end{tabular}
    }%
    \vspace{3ex}
    \subcaptionbox{(b) Videos \& Duration per Year\label{tab:dist_year}}[\linewidth]{%
        \begin{tabular}{r rr r}
          \toprule
          Year & Count & Hours & Share \\
          \midrule
          2000 & 50 & 12.3 & 16.6\,\% \\
          2004 & 26 & 12.5 & 16.9\,\% \\
          2006 & 20 &  8.3 & 11.2\,\% \\
          2008 & 75 & 16.2 & 21.9\,\% \\
          2012 & 35 & 21.5 & 29.1\,\% \\
          2013 & 20 &  3.1 &  4.3\,\% \\
          \midrule
          All  & 226 & 74.0 & 100.0\,\% \\
          \bottomrule
        \end{tabular}
    }%
  \end{minipage}
  \hfill
  \begin{minipage}[t]{0.32\textwidth}
    \subcaptionbox{(c) Video FPS Distribution\label{tab:dist_video_fps}}[\linewidth]{%
      \begin{tabular}{rrr}
        \toprule
        FPS & Count & Share \\
        \midrule
        29.97 & 192 & 85.0\,\% \\
        25.00 &  34 & 15.0\,\% \\
        \bottomrule
      \end{tabular}
    }%
    \vspace{4.3ex}
    \subcaptionbox{(d) Audio Sample Rate Distribution\label{tab:dist_audio_sr}}[\linewidth]{%
      \begin{tabular}{rrr}
        \toprule
        Rate (Hz) & Count & Share \\
        \midrule
        48000  & 150 & 66.4\,\% \\
        32000  &  76 & 33.6\,\% \\
        \bottomrule
      \end{tabular}
    }%
    \vspace{4.2ex}
    \subcaptionbox{(e) Audio Channels Distribution\label{tab:dist_audio_ch}}[\linewidth]{%
      \begin{tabular}{crr}
        \toprule
        Channels & Count & Share \\
        \midrule
        2 & 205 & 90.7\,\% \\
        6 &  21 &  9.3\,\% \\
        \bottomrule
      \end{tabular}
    }%
  \end{minipage}
  \hfill
  \begin{minipage}[t]{0.30\textwidth}
    \subcaptionbox{(f) Video Duration Histogram\label{fig:dist_placeholder}}[\linewidth]{%
      \centering
      \includegraphics[width=1\textwidth]{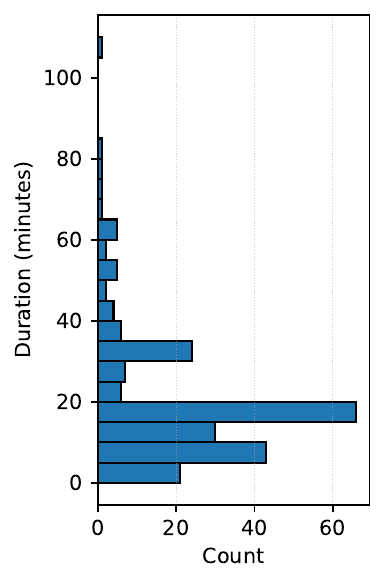}
    }%
  \end{minipage}

  \caption{\dataset{70h}: Media attribute distributions.}
  \label{tab:bc70h_stats}
\end{table}

\end{document}